\documentclass{article}
\pdfoutput=1

\usepackage[preprint]{neurips_2026}

\usepackage[utf8]{inputenc} % allow utf-8 input
\usepackage[T1]{fontenc}    % use 8-bit T1 fonts
\usepackage{hyperref}       % hyperlinks
\usepackage{url}            % simple URL typesetting
\usepackage{booktabs}       % professional-quality tables
\usepackage{amsfonts}       % blackboard math symbols
\usepackage{nicefrac}       % compact symbols for 1/2, etc.
\usepackage{microtype}      % microtypography
\usepackage{xcolor}         % colors
\usepackage{graphicx}
\usepackage{twemojis}
\usepackage{subcaption}
\usepackage{multirow}
\usepackage{amsmath}
\usepackage{amssymb}
\usepackage{mathtools}
\usepackage{amsthm}
\usepackage[capitalize,noabbrev]{cleveref}

\usepackage{tikz}
\usepackage{pgfplots}
\pgfplotsset{compat=1.17}
\usetikzlibrary{plotmarks}

\definecolor{c1}{RGB}{0,114,178}   
\definecolor{c2}{RGB}{230,159,0}   
\definecolor{c3}{RGB}{0,158,115}   
\definecolor{c4}{RGB}{204,121,167} 
\definecolor{c5}{RGB}{213,94,0}

\theoremstyle{plain}

\theoremstyle{definition}

\theoremstyle{remark}

\title{Prune, Interpret, Evaluate (PIE \scalebox{2.5}{\twemoji{pie}}): A Cross-Layer Transcoder-Native Framework for Efficient Circuit Discovery via Feature Attribution}

\author{%
  Qinhao Chen \\
  Columbia University \\
 \\
  \And
  Linyang He\thanks{Corresponding author.} \\
  Columbia University \\
  \texttt{linyang.he@columbia.edu} \\
  \And
  Nima Mesgarani \\
  Columbia University \\
 \\
}

\begin{document}

\maketitle

\begin{abstract}
  Existing feature-interpretation pipelines typically operate on uniformly sampled units or exhaustive feature sets, incurring massive costs on units irrelevant to target behaviors. To address this, we introduce the first CLT-native end-to-end pruning framework, \textbf{PIE}, which pioneers the paradigm of pruning first and interpreting later. PIE connects Pruning, automatic Interpretation, and interpretation Evaluation, establishing a comprehensive benchmarking environment to systematically measure behavioral fidelity and downstream interpretability under pruning. Within this framework, we adapt strong relevance baselines and propose Feature Attribution Patching (FAP), a patch-grounded attribution method that scores CLT features by aggregating gradient-weighted write contributions. Furthermore, we introduce FAP-Synergy, a systematic synergy-aware reranking procedure. We evaluate pruning using KL-divergence and Faithfulness behavior retention and assess interpretation quality with FADE-style metrics across IOI and Doc-String tasks. Across budget constraints of $K\in\{50,100,200,400,800\}$, our rigorous benchmarking reveals distinct operational regimes: while base FAP and adapted baselines perform more efficiently at relaxed budgets, FAP-Synergy excels in strict-budget regimes. Crucially, we demonstrate a practical ``Effective Budget'' advantage: on the IOI task for both Llama-3.2-1B and Gemma-2-2B, FAP-Synergy at $K{=}50$ functionally matches the behavioral fidelity of baseline circuits at $K{=}75$. Because downstream evaluation costs scale linearly per feature, Synergy effectively grants the pipeline 25 ``free'' features, achieving $K{=}75$ fidelity while reducing interpretation costs by 33\%. Our code is available at \url{https://github.com/Qinhao-Chen/PIE-Pipeline}.
\end{abstract}

\section{Introduction}

Mechanistic interpretability seeks to explain model behavior by identifying internal components and causal pathways mediating inputs to outputs \citep{Olah2022Mechanistic}. This is increasingly urgent as large language models (LLMs) enter high-stakes domains demanding transparency and targeted intervention (e.g., safety and biomedical support) \citep{templeton2024scaling,Yang2022UnboxBlackBox,Band2023XAIHealth,chen2025safetyneurons}. Recent progress has advanced from neuron-level anecdotes to scalable \emph{feature- and circuit-level} accounts, driven by sparse autoencoders (SAEs) and replacement-model approaches like cross-layer transcoders (CLTs) \citep{Bricken2023Monosemanticity,Ameisen2025CircuitTracing,Conmy2023ACDC,Syed2024EAP}.
\begin{figure*}[t]
    \centering
    \includegraphics[width=1.0\textwidth]{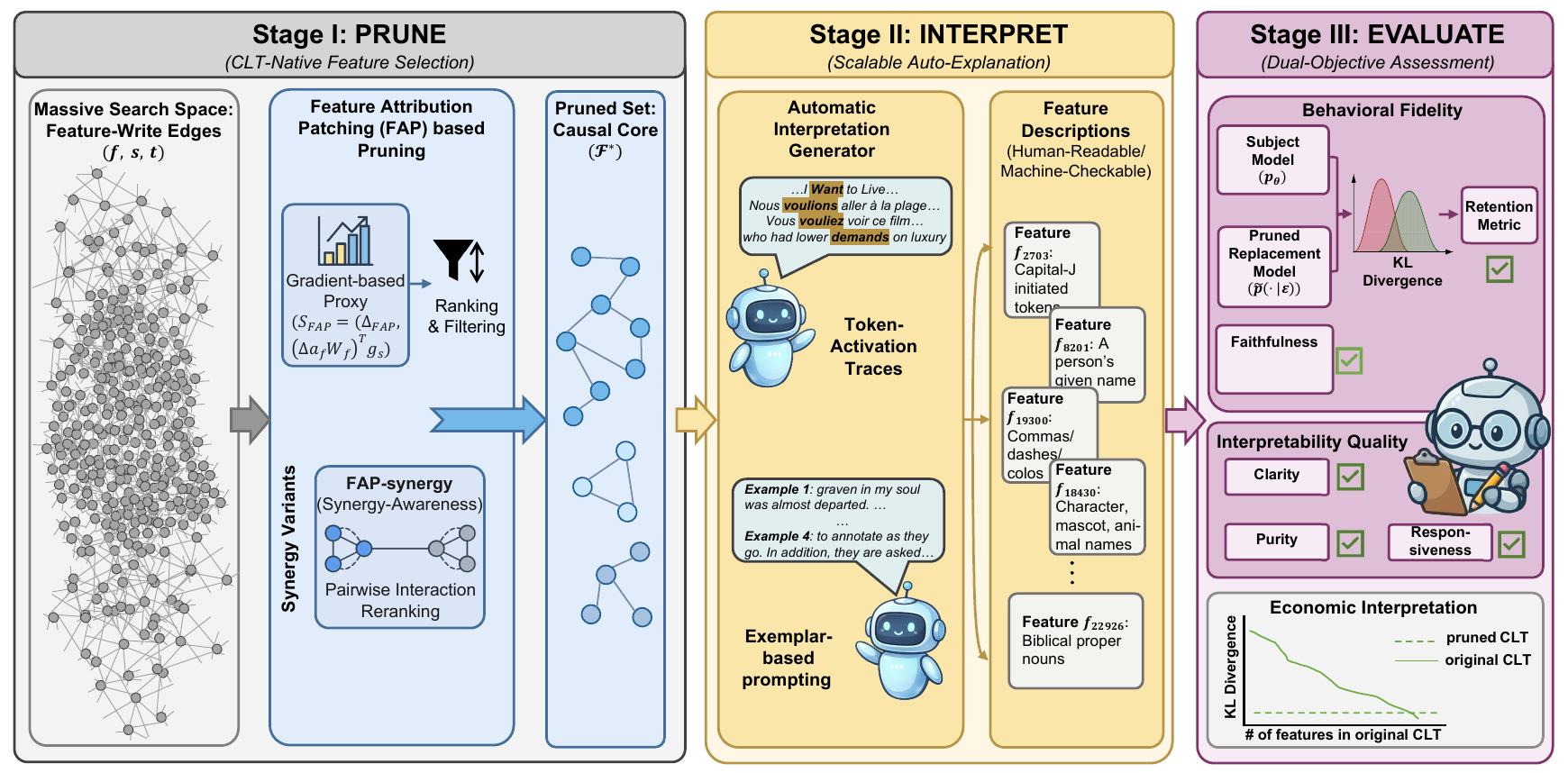}
    
    \caption{The PIE Framework: A CLT-Native End-to-End Pipeline. \textbf{Stage I (Prune)} filters the massive search space of CLT feature-write edges into a sparse "Causal Core" using Feature Attribution Patching (FAP) and its synergy-aware variant (FAP-Synergy), which systematically reranks boundary features via pairwise interactions. \textbf{Stage II (Interpret)} generates natural language descriptions only for retained features using exemplar-based prompting, drastically reducing costs. \textbf{Stage III (Evaluate)} performs a dual-objective assessment: quantifying behavioral fidelity (KL Divergence, PCR) and measuring interpretation quality via automated metrics (Clarity, Purity, Responsiveness).}
    \label{fig:prune_pipeline}
\end{figure*}

Despite these advances, a central bottleneck remains: \textit{allocating the interpretation budget}. Modern dictionaries contain millions of features, making downstream auto-interpretation and evaluation expensive per feature \citep{Bills2023NeuronExplainer,Paulo2024MillionsFeatures,puri2025fade,Boggust2025SemanticRegexes}. End-to-end workflows must therefore allocate compute carefully; explaining \emph{everything} is infeasible, and explaining uninformative features wastes resources.

\paragraph{Motif 1: Pruning is necessary, and CLTs demand a native interpretability framework.}
Existing pipelines often evaluate uniformly sampled features, creating redundancy by analyzing weakly causal units \citep{Bills2023NeuronExplainer,Paulo2024MillionsFeatures,puri2025fade,Boggust2025SemanticRegexes}. This motivates our principle: \emph{prune first, interpret later}. Furthermore, because CLT replacement models utilize cross-layer feature writes not naturally captured by SAE-only pipelines \citep{Ameisen2025CircuitTracing,lindsey2025landscape}, scalable interpretability requires a CLT-native framework. PIE achieves this by establishing a comprehensive benchmarking environment that connects pruning, interpretation, and evaluation to systematically measure how pruning impacts fidelity and explainability.

\paragraph{Motif 2: Mainstream pruning ignores synergy.}
Standard pruning assumes additive importance, selecting high-magnitude edges and discarding the rest \citep{Syed2024EAP,hanna2024faithfaithfulnessgoingcircuit}. However, CLT circuits violate this through \emph{non-additive interactions} (``synergy''), where individually weak components matter jointly—effects we empirically verify (Sec.~\ref{sec:experiments_results}). Ignoring synergy systematically underestimates crucial components, yielding behaviorally degraded circuits.

\paragraph{Our approach: Feature Attribution Patching (FAP).} 
We introduce a CLT-native pruning family operating directly on \emph{features}. Our base method, FAP, scores occurrences using a fast, patch-grounded first-order estimate derived from cross-layer write differences and cached downstream gradients. To account for non-additive interactions, we introduce FAP-Synergy, a variant that rescues near-threshold features based on pairwise complementarity with the selected core. While developed as a general methodological enhancement, we empirically demonstrate that this interaction-awareness becomes highly advantageous in strictly constrained budget regimes.

This paper makes four main contributions:
\begin{itemize}
    \item \textbf{Prune-First Interpretability Framework:} We present \textbf{PIE}, the first end-to-end framework connecting CLT-native pruning, automatic interpretation, and evaluation, shifting the paradigm from exhaustive interpretation to systematic, benchmarked extraction.
    \item \textbf{Feature Attribution Patching (FAP):} We propose an efficient, gradient-weighted patching method that scores sparse cross-layer feature writes, enabling highly scalable pruning.
    \item \textbf{FAP-Synergy \& Regime-Specific Insights:} We introduce a synergy-aware reranking method. Our benchmarking reveals that while base FAP suffices for relaxed budgets, FAP-Synergy excels under strict constraints, matching baselines' fidelity at $K{=}75$ using only $K{=}50$ features—a 33\% reduction in downstream costs.
    \item \textbf{Rigorous Baselines \& Evaluation:} We adapt Relevance Patching (\textbf{CLT-RelP}) into a CLT-native format to serve as a strong baseline, and systematically quantify pruning quality using FADE-style interpretability metrics.

\end{itemize}

\section{Related Work}
\label{sec:related}

\paragraph{Attribution patching and edge pruning.}
Causal subgraph search builds on network pruning, historically estimating unit relevance via first-order error derivatives \citep{mozer1988skeletonization,molchanov2017pruningconvolutionalneuralnetworks} or expensive second-order Hessians \citep{lecun1989OBD,hassibi1993OBS}. Modern circuit discovery adapts this to computation graphs, shifting from iterative evaluations \citep{Conmy2023ACDC} to efficient gradient-based proxies like Edge Attribution Patching (EAP) \citep{Syed2024EAP} and Relevance Patching (RelP) \citep{jafari2025relpfaithfulefficientcircuit}. We inherit this patching-first lineage but shift the unit of analysis: instead of pruning module edges or weights, we prune CLT feature writes. While second-order methods capture complex interactions, Hessian computation scales quadratically ($O(N^2)$). Thus, FAP uses a fast first-order Taylor approximation to score cross-layer writes, enabling scalable, feature-level pruning. For rigorous benchmarking, we adapt RelP—originally restricted to standard modules—into a CLT-native format (\textbf{CLT-RelP}), providing a competitive baseline.

\paragraph{Feature-based interpretability and CLT-native circuit tracing.}
Because individual neurons are often polysemantic, mechanistic interpretability increasingly relies on learned sparse feature bases as stable semantic units \citep{Bricken2023Monosemanticity}. Superposition toy models and related phenomena suggest feature directions, rather than neurons, are the minimal units for interpretation \citep{elhage2022toymodels,henighan2023doubledescent,power2022grokking}. Replacement models enable causal validation via controlled interventions on feature flow. Specifically, CLT circuit tracing represents computation as sparse cross-layer writes, supporting causal experiments where features are patched, ablated, or amplified \citep{Ameisen2025CircuitTracing,dunefsky2024transcoders}. Resources like Neuronpedia’s circuit graphs operationalize this, highlighting the need for methods operating directly on CLT-native objects \citep{lindsey2025landscape}. Our method explicitly targets this setting, scoring and pruning CLT features by aggregating cross-layer writes rather than reverting to neuron- or SAE-only views.

\paragraph{Automated interpretation and evaluation of feature descriptions.}
Complementary work reduces human effort via automated interpretation of internal units. Explainer systems prompt LLMs with activation traces or exemplars to generate descriptions of unit representations \citep{Bills2023NeuronExplainer,Paulo2024MillionsFeatures,Lin2023Neuronpedia}. To further refine description quality, some approaches tie hypotheses directly to behavior \citep{GurArieh2025OutputCentric}, while others explore structured, executable formats such as semantic regex-style DSLs \citep{Huang2023RigorousNeuronExplanations,Boggust2025SemanticRegexes}. Concurrently, frameworks like FADE formalize interpretation quality via metrics (clarity, purity, responsiveness) and causal validation \citep{puri2025fade}. While advancing description generation and evaluation, these systems assume a large candidate pool, ignoring the upstream bottleneck: \emph{which features should be explained}.

\section{Method: The PIE Framework}
\label{sec:method}

We introduce \textbf{PIE} (Prune, Interpret, Evaluate), a CLT-native framework designed to resolve the efficiency bottleneck in mechanistic interpretability. While SAE features offer semantic sparsity, interpreting millions of features is computationally prohibitive. Our framework rests on the premise that \emph{pruning must precede interpretation}: by identifying the sparse subset of Cross-Layer Transcoder (CLT) features causally relevant to a task, we can allocate interpretation budgets where they matter most.

The framework consists of three stages: (1) \textbf{Pruning} via Feature Attribution Patching (FAP) and its variants to select a minimal, high-fidelity circuit; (2) \textbf{Automatic Interpretation} of the retained features; and (3) \textbf{Evaluation} of both behavioral fidelity and explanation quality.

\subsection{Problem Setup: CLT-Native Features}
\label{sec:clt_setup}

We analyze a subject model $M$ utilizing a Cross-Layer Transcoder (CLT) as a replacement model. Unlike standard transformers where edges connect modules (e.g., Attention Head $\to$ MLP), a CLT decomposes computation into sparse feature activations that write directly across layers.

Let a feature $f$ in the CLT dictionary have activation $a_f(x, t)$ at token position $t$ for input $x$. This feature contributes a vector $W^{(s)}_f \in \mathbb{R}^{d_{model}}$ to a downstream residual stream site $s$. We define the fundamental unit of analysis as the \textbf{feature-write edge}: the specific contribution of feature $f$ to site $s$ at position $t$.
Pruning involves selecting a subset of features $\mathcal{F}^*$ (where $|\mathcal{F}^*| \ll |\mathcal{F}_{total}|$) such that the replacement model maintains low KL-divergence from the original subject model $M$.

\subsection{Pruning via Feature Attribution Patching (FAP)}
\label{sec:fap}

To scale pruning to millions of features, we propose \textbf{Feature Attribution Patching (FAP)}, a first-order approximation method adapted for CLT feature space.

\paragraph{Gradient-Weighted Write Attribution.}
We estimate feature importance by combining activation differences with downstream gradients. Given a clean input $x_{\text{clean}}$ and a corrupted input $x_{\text{corr}}$, let $\Delta a_f(t) = a_f(x_{\text{clean}}, t) - a_f(x_{\text{corr}}, t)$ be the activation difference for feature $f$ at position $t$.
We compute the gradient of a target metric $\mathcal{L}$ (e.g., logit difference or negative KL) with respect to the receiver site activation $h_s(t)$, denoted as $\nabla_{h_s(t)} \mathcal{L}$. 

Instead of collapsing positional data into a global feature score, the FAP score for a specific feature occurrence $(f, t)$ aggregates its downstream impact across all target receiver sites $s$:
\begin{equation}
    S_{\text{FAP}}(f, t) = \sum_{s} \left( \Delta a_f(t) \cdot W^{(s)}_f \right)^\top \nabla_{h_s(t)} \mathcal{L}
\end{equation}
This score efficiently estimates how much the specific occurrence of feature $f$ at position $t$ contributes to restoring the clean behavior. For each input prompt, we apply a strict \textbf{per-prompt feature occurrence budget}, retaining only the top-$K$ feature occurrences with the highest magnitude scores $|S_{\text{FAP}}(f, t)|$. 

To maintain precise bookkeeping between pruning and interpretation: the budget $K$ applies locally to \textit{occurrences} within a single forward pass. However, our downstream automatic interpretation phase operates on \textit{unique features}. Therefore, the final interpretation budget is determined by the size of the global unique feature set, $\mathcal{F}^*_{\text{unique}}$, which is the union of all retained feature occurrences across the entire dataset.

\subsection{Addressing Attribution Selection Failure Modes}
\label{sec:synergy}

While vanilla FAP is efficient, selecting features purely by \textbf{attribution-magnitude} (i.e., $|S_{FAP}(f)|$) suffers from \emph{synergy ignorance} (missing features that only matter deeply in combination). We introduce a specialized variant to address these.

\paragraph{FAP-Synergy: Boundary Interaction Reranking.}
Attribution-magnitude selection often fails at the "pruning boundary"---the threshold separating kept and pruned features. Features just below the cutoff may be individually weak but highly synergistic with features already selected. To systematically integrate these interaction signals, we partition candidates into a confident "Core" ($\mathcal{F}_{\text{core}}$) and a marginal "Boundary" ($\mathcal{F}_{\text{bound}}$). We standardize a boundary feature's base attribution, $z_{\text{base}}(f_b)$, relative to the median core magnitude. To ensure scalable evaluation, we estimate its normalized synergy, $z_{\text{syn}}(f_b)$, by sampling a subset of core partners ($\mathcal{P} \subset \mathcal{F}_{\text{core}}$, $|\mathcal{P}|=8$), computing the median pairwise synergy:
\begin{equation}
    \text{Syn}(f_b, f_c) = \mathcal{M}(\{f_b, f_c\}) - \mathcal{M}(\{f_b\}) - \mathcal{M}(\{f_c\})
\end{equation}
where $\mathcal{M}$ represents the metric recovery relative to the baseline. We then divide this median by the partners' median individual effects. Crucially, we isolate constructive complementarity by restricting synergy to the positive domain, $z_{\text{syn}}^+ = \max(0, z_{\text{syn}})$. This ensures redundant features are not artificially penalized below their base gradient attribution. The final systematic reranking score is $S'_{f_b} = z_{\text{base}}(f_b) + \lambda \cdot z_{\text{syn}}^+(f_b)$, which strictly determines the exact $K$-feature retention, aiming to lower the KL divergence without increasing the feature budget.

\subsection{Automatic Interpretation \& Evaluation}
\label{sec:eval_pipeline}

Once features are pruned, we generate natural language explanations using an LLM (e.g., GPT-5.2) prompted with max-activating exemplars \citep{Bills2023NeuronExplainer}. Crucially, because we interpret only the pruned set, we drastically reduce API costs compared to dense sweeps. We evaluate the quality of the discovery process using two categories of metrics.

\paragraph{Behavioral Fidelity Metrics.}
We measure how well the pruned feature set $\mathcal{F}^*$ preserves the original model mechanics using two metrics: \textbf{KL Divergence}, measured at the last token between the subject model's output distribution and the replacement model restricted to $\mathcal{F}^*$ (lower is better); and \textbf{Faithfulness} \citep{jafari2025relpfaithfulefficientcircuit}. Faithfulness enables direct and fair comparison with contemporary relevance-propagation methods (e.g., our adapted CLT-RelP) by measuring the proportion of original performance captured by the discovered circuit. Formally, it is calculated as $\text{Faithfulness} = (\mathcal{L}(C) - \mathcal{L}(\emptyset)) / (\mathcal{L}(\mathcal{M}) - \mathcal{L}(\emptyset))$, where $\mathcal{L}(C)$ is the logit difference evaluated using only the active pruned circuit $C$, $\mathcal{L}(\emptyset)$ is the logit difference of the fully corrupted baseline, and $\mathcal{L}(\mathcal{M})$ is the logit difference of the full, unablated model. An ideal circuit contains all necessary mechanisms to perfectly recover the output preference, yielding a score of $1.0$.

\paragraph{Interpretability Metrics.}
To ensure that FAP selects meaningfully interpretable units (rather than just polysemantic error-correcting terms), we evaluate the generated explanations using the FADE framework metrics \citep{puri2025fade}: \textbf{Clarity}, assessing if an independent auditor LLM can generate synthetic samples that activate the feature based \emph{only} on the description (measured via Gini coefficient of activations on synthetic vs. control data); \textbf{Purity}, checking if the explanation can distinguish high-activating from low-activating real dataset examples (measured via Average Precision); and \textbf{Responsiveness}, measuring whether natural samples aligning with the description consistently trigger the feature (measured via Gini coefficient on rated natural samples).

\section{Experiments and Results}
\label{sec:experiments_results}

We evaluate the PIE framework on Llama-3.2-1B \citep{grattafiori2024llama3herdmodels} and Gemma-2-2B \citep{gemmateam2024gemma2improvingopen} utilizing pre-trained Cross-Layer Transcoders (CLTs) \citep{circuit-tracer}. To validate our approach, we employ a distribution-split protocol: we extract and prune circuits on the Indirect Object Identification (IOI) task \citep{wang2023interpretability} and Doc-String task \citep{Heimersheim2023Docstrings} across feature budgets $K\in\{50,100,200,400,800\}$, and subsequently evaluate explanation quality based on their activations on a held-out dataset of 2 million Wikipedia sentences \citep{sentence-transformers_wikipedia-en-sentences_2024}. We benchmark our gradient-based \textbf{FAP} and boundary-reranking \textbf{FAP-Synergy} ($\lambda=3$, top-25\% boundary) methods against three baselines: \textbf{Activation-Magnitude} pruning, perturbation-based \textbf{Feature Activation Patching} (FActP, ACDC-style pruning), and our novel CLT adaptation of \textbf{Relevance Patching (CLT-RelP)}, which replaces local gradients with Layer-wise Relevance Propagation (LRP) coefficients for faithful circuit discovery over features rather than standard modules \citep{jafari2025relpfaithfulefficientcircuit}. Detailed model, dataset, and hyperparameter configurations are provided in Appendices~\ref{app:hyperparam} and~\ref{app:implementation}.

\subsection{Behavioral Fidelity and Pruning Efficiency}
\label{sec:fidelity_results}

We evaluate pruning quality across two tasks, five budgets ($K \in \{50, 100, 200, 400, 800\}$), and five feature-selection methods. Our analysis deliberately partitions the evaluation into distinct operational regimes. Due to space constraints, we focus our main-text analysis (Tables~\ref{tab:kl_sweeps_main_short} and \ref{tab:faithfulness_sweeps_short}) exclusively on highly constrained, interpretable regimes ($K \le 200$). This focus is driven by two benchmarking principles:

First, human comprehension---the core objective of mechanistic interpretability---deteriorates at larger budgets (e.g., 400+ features). Thus, evaluating pruning methods in these relaxed regimes is less relevant to the practical bottlenecks of interpretability workflows.

Second, as detailed in our extended evaluations (Appendix~\ref{sec:appendix_extended_fidelity}), gradient and patching methods empirically converge at large budgets. Because precise feature ranking near the boundary becomes less critical when the budget captures the vast majority of meaningful features, we reserve the extended results for the Appendix. Therefore, our main text highlights the strict-to-medium budget regimes ($K \le 200$) where algorithmic choice drastically impacts fidelity. In these constrained settings, the FAP family consistently achieves the lowest KL, with FAP-Synergy providing the most significant advantages.

\begin{table*}[ht]
\centering
\caption{\textbf{Behavioral fidelity across tasks for interpretable budgets ($K \le 200$).} We report mean last-token KL (lower is better). The FAP family consistently outperforms Activation-Magnitude. FAP-Synergy yields the clearest gains in the strictest regimes. Results for extended budgets are in Appendix~\ref{sec:appendix_extended_fidelity}.}
\label{tab:kl_sweeps_main_short}
\small
\setlength{\tabcolsep}{5pt}
\begin{tabular}{llccc}
\toprule
\textbf{Task / Model} & \textbf{Method} & \textbf{$K{=}50$} & \textbf{$K{=}100$} & \textbf{$K{=}200$} \\
\midrule
\multicolumn{5}{l}{\textit{IOI --- Llama-3.2-1B}} \\
& CLT-RelP & $1.32 \pm 0.62$ & $1.15 \pm 0.54$ & $0.87 \pm 0.41$ \\
& FAP & $1.33 \pm 0.60$ & $1.13 \pm 0.53$ & $\mathbf{0.85 \pm 0.42}$ \\
& FAP-Synergy & $\mathbf{1.22 \pm 0.56}$ & $\mathbf{1.12 \pm 0.52}$ & $\mathbf{0.85 \pm 0.42}$ \\
& Activation-Magnitude & 1.59 $\pm$ 0.69 & 1.52 $\pm$ 0.65 & 1.32 $\pm$ 0.59 \\
& FActP & $1.26 \pm 0.63$ & $1.24 \pm 0.64$ & $1.22 \pm 0.63$ \\
\midrule
\multicolumn{5}{l}{\textit{IOI --- Gemma-2-2B}} \\
& CLT-RelP & $0.86 \pm 0.43$ & $0.75 \pm 0.37$ & $0.53 \pm 0.28$ \\
& FAP & $0.91 \pm 0.46$ & $0.74 \pm 0.37$ & $\mathbf{0.52 \pm 0.29}$ \\
& FAP-Synergy & $0.82 \pm 0.42$ & $\mathbf{0.73 \pm 0.36}$ & $\mathbf{0.52 \pm 0.29}$ \\
& Activation-Magnitude & 1.29 $\pm$ 0.59 & 1.25 $\pm$ 0.58 & 1.18 $\pm$ 0.56 \\
& FActP & $0.81 \pm 0.37$ & $0.77 \pm 0.38$ & $0.76 \pm 0.37$ \\
\midrule
\multicolumn{5}{l}{\textit{Doc-String --- Llama-3.2-1B}} \\
& CLT-RelP & $0.80 \pm 0.16$ & $0.72 \pm 0.15$ & $0.55 \pm 0.12$ \\
& FAP & $0.78 \pm 0.15$ & $0.69 \pm 0.14$ & $\mathbf{0.53 \pm 0.12}$ \\
& FAP-Synergy & $\mathbf{0.74 \pm 0.14}$ & $\mathbf{0.68 \pm 0.14}$ & $\mathbf{0.53 \pm 0.12}$ \\
& Activation-Magnitude & 0.88 $\pm$ 0.17 & 0.88 $\pm$ 0.17 & 0.85 $\pm$ 0.16 \\
& FActP & $0.78 \pm 0.16$ & $0.76 \pm 0.15$ & $0.76 \pm 0.15$ \\
\midrule
\multicolumn{5}{l}{\textit{Doc-String --- Gemma-2-2B}} \\
& CLT-RelP & $0.42 \pm 0.11$ & $0.37 \pm 0.10$ & $0.31 \pm 0.09$ \\
& FAP & $0.41 \pm 0.11$ & $\mathbf{0.36 \pm 0.10}$ & $0.30 \pm 0.09$ \\
& FAP-Synergy & $\mathbf{0.40 \pm 0.11}$ & $\mathbf{0.36 \pm 0.10}$ & $\mathbf{0.29 \pm 0.09}$ \\
& Activation-Magnitude & 0.43 $\pm$ 0.12 & 0.43 $\pm$ 0.12 & 0.43 $\pm$ 0.12 \\
& FActP & $0.40 \pm 0.12$ & $0.40 \pm 0.12$ & $0.39 \pm 0.12$ \\
\bottomrule
\end{tabular}
\end{table*}

\begin{table*}[ht]
\centering
\caption{\textbf{Faithfulness on IOI across interpretable budgets.} We report mean faithfulness $\pm$ standard deviation. (Higher is better, though variance is also shown). Extended budgets are in Appendix~\ref{sec:appendix_extended_fidelity}.}
\label{tab:faithfulness_sweeps_short}
\small
\setlength{\tabcolsep}{5pt}
\begin{tabular}{llccc}
\toprule
\textbf{Task / Model} & \textbf{Method} & \textbf{$K{=}50$} & \textbf{$K{=}100$} & \textbf{$K{=}200$} \\
\midrule
\multicolumn{5}{l}{\textit{IOI --- Llama-3.2-1B}} \\
& CLT-RelP & $0.15 \pm 0.17$ & $0.22 \pm 0.20$ & $0.37 \pm 0.22$ \\
& FAP & $0.27 \pm 0.17$ & $0.34 \pm 0.17$ & $0.43 \pm 0.17$ \\
& FAP-Synergy & $0.30 \pm 0.16$ & $0.35 \pm 0.15$ & $0.43 \pm 0.16$ \\
\midrule
\multicolumn{5}{l}{\textit{IOI --- Gemma-2-2B}} \\
& CLT-RelP & $0.26 \pm 0.45$ & $0.35 \pm 0.60$ & $0.54 \pm 0.54$ \\
& FAP & $0.29 \pm 0.38$ & $0.39 \pm 0.43$ & $0.55 \pm 0.46$ \\
& FAP-Synergy & $0.34 \pm 0.40$ & $0.40 \pm 0.44$ & $0.55 \pm 0.46$ \\
\bottomrule
\end{tabular}
\end{table*}

\paragraph{Method comparison \& The Failure of FActP.}
Activation-Magnitude pruning is consistently weaker than the FAP family across both tasks, indicating that clean activation size alone cannot isolate the causal core. FActP performs competitively only at $K{=}50$, but severely plateaus. Upon inspecting the underlying ACDC process, we found that CLTs are highly robust to single-feature ablations due to feature redundancy and compensatory mechanisms. Consequently, substituting individual features often yields zero effect. We observed a low number of features with non-zero ACDC effects (averaging 185.18 for Gemma-2-2B and 254.54 for Llama-3.2-1B). Because it fails to rank the medium-importance tail, FActP degrades to random selection past $K \approx 200$. Conversely, FAP's gradient estimation consistently and accurately ranks medium-to-low importance features.

\paragraph{Comparison with CLT-RelP.} 
Our adapted Relevance Patching (CLT-RelP) serves as a competitive baseline, generally outperforming magnitude and single-feature methods. However, in the strict-budget regimes necessary for human interpretability ($K \le 100$), the FAP family demonstrates distinct advantages, underscoring its design for highly constrained settings. In terms of behavioral fidelity (Table~\ref{tab:kl_sweeps_main_short}), FAP-Synergy consistently edges out CLT-RelP at $K=50$ (e.g., $1.22$ vs. $1.32$ on Llama). This is more pronounced in Faithfulness (Table~\ref{tab:faithfulness_sweeps_short}), where base FAP on Llama at $K=50$ achieves double the faithfulness of CLT-RelP ($0.30$ vs. $0.15$). On Gemma-2-2B, CLT-RelP suffers from high variance, whereas the FAP family maintains monotonic, low-variance improvement.

\paragraph{Budget dependence and convergence at scale.}
FAP-Synergy is most crucial in low-budget regimes ($K{=}50,100$), where selection near the pruning boundary is highly competitive. For example, on IOI at $K{=}50$, FAP-Synergy reduces KL from 1.33 to 1.22 on Llama and from 0.91 to 0.82 on Gemma. As shown in Appendix~\ref{sec:appendix_extended_fidelity}, as the budget scales past $K \ge 400$, CLT-RelP, base FAP, and FAP-Synergy effectively converge. This is expected: a sufficiently large budget set will naturally encapsulate most meaningful features, rendering the precise ranking of the 400th versus 401st feature largely irrelevant. Therefore, for analysis pipelines operating under relaxed sparsity constraints, we explicitly recommend deploying base FAP or CLT-RelP to maximize computational efficiency.

\paragraph{The "Effective Budget" (Cost Equivalence).}
To directly quantify the efficiency gains of FAP-Synergy in tight-budget regimes, we report a supplementary baseline experiment at $K{=}75$ on the IOI task. For Llama-3.2-1B, base FAP and CLT-RelP at $K{=}75$ yield KL divergences of $1.22$ and $1.23$, respectively; FAP-Synergy matches this at $K{=}50$ with a KL of $1.22$. Similarly, for Gemma-2-2B, base FAP ($0.81$) and CLT-RelP ($0.80$) at $K{=}75$ are functionally matched by FAP-Synergy at $K{=}50$ ($0.82$). Because downstream evaluation costs scale linearly per feature, Synergy effectively grants the pipeline 25 ``free'' feature occurances, achieving $K{=}75$ fidelity while reducing interpretation costs by 33\%.

\paragraph{Compression relative to active-feature random selection.}
Figure~\ref{fig:fidelity_curve} compares PIE against random selection from the prompt-active feature set. This measures how many active features are needed for random selection to match the fidelity achieved by PIE at $K{=}100$. In both models, PIE reaches a fidelity level that random active-set selection only achieves at roughly $4\text{k}$ features, yielding an approximate $40\times$ compression.

\begin{figure}[ht]
    \centering
    \begin{tikzpicture}
    \begin{axis}[
        width=0.48\textwidth,
        height=0.35\textwidth,
        xlabel={Number of Features Kept ($K$)},
        ylabel={KL Divergence (nats)},
        grid=both,
        major grid style={line width=.2pt,draw=gray!50},
        legend style={at={(0.95,0.95)},anchor=north east, font=\tiny},
        tick label style={font=\tiny},
        label style={font=\scriptsize},
        xmin=0, xmax=5200,
        ymin=0, ymax=22,
    ]
    
    \addplot[color=c1, mark=*, mark size=1.5pt] coordinates {
        (100, 19.73)
        (1000, 15.25)
        (2500, 6.59)
        (4000, 1.03)
        (4250, 0.69)
        (4500, 0.42)
    };
    \addlegendentry{Gemma Random}

    \addplot[color=c1, dashed, thick] coordinates {(0, 0.74) (5200, 0.74)};
    \addlegendentry{Gemma FAP ($K{=}100$)}

    \addplot[color=c5, mark=square*, mark size=1.5pt] coordinates {
        (100, 12.30)
        (1000, 11.36)
        (2000, 9.07)
        (3000, 3.88)
        (3500, 1.77)
        (3750, 1.15)
        (4000, 0.67)
        (4250, 0.35)
    };
    \addlegendentry{Llama Random}

    \addplot[color=c5, dashed, thick] coordinates {(0, 1.128) (5200, 1.128)};
    \addlegendentry{Llama FAP ($K{=}100$)}

    \end{axis}
    \end{tikzpicture}
    \caption{\textbf{The Causal Sparsity Gap.} We plot the KL divergence of randomly sampling from the prompt-active feature set as budget $K$ increases (solid lines). The dashed lines represent the KL achieved by PIE's causal extraction (FAP) at a strict budget of $K{=}100$. Random sampling requires $\approx$4,000 features to match PIE's fidelity. This massive $\approx$40$\times$ gap demonstrates that the vast majority of active features are causally irrelevant, strongly motivating the necessity of a prune-first paradigm prior to auto-interpretation.}
    \label{fig:fidelity_curve}
\end{figure}

\subsection{The Signal-to-Noise Interpretability Gap}

A primary hypothesis of this work is that task-relevant features are inherently more interpretable than the general population. We quantify this by measuring the distribution of interpretation quality metrics (Clarity, Purity, Responsiveness) on Docstring tasks over budget $K=50$ and $K=100$ (as the total covered features grow combinatorially with the budget, evaluating larger sets becomes prohibitively expensive). 

As shown in Table~\ref{tab:interpretability_results}, FAP-based pruning strategies demonstrate strong performance against relevance-based (RelP) and factorization-based (FActP) methods. Crucially, at the tightest bottleneck of $K=50$, FAP-Synergy establishes a clear advantage, yielding the highest clarity and purity. This confirms that incorporating boundary interactions systematically rescues features with high semantic distinctness and conceptual coherence. While FAP and FAP-Synergy maintain competitive quality at $K=100$, the metrics across all methods begin to compress, further validating that synergy-awareness is most critical when the interpretation budget is strictly limited.

Due to space limitations, the full interpretability results for the IOI task are deferred to Appendix~\ref{app:fade_ioi}. However, the empirical trends remain highly consistent: at a strict $K=50$ budget, FAP-Synergy yields the highest overall Clarity and Purity scores across both Llama-3.2-1B and Gemma-2-2B models, demonstrating the cross-task robustness of synergy-aware feature selection in constrained regimes.

\begin{table*}[ht]
\centering
\caption{Interpretability quality of pruned circuits at $K=50$ and $K=100$ budgets on Doc-String tasks. We report Mean $\pm$ Std for quality metrics. FAP-Synergy maintains the highest overall interpretability.}
\label{tab:interpretability_results}
\begin{small}
\begin{sc}
\begin{tabular}{lccc}
\toprule
\textbf{Method} & \textbf{Clarity}  & \textbf{Purity}  & \textbf{Responsiveness}  \\
\midrule
\multicolumn{4}{l}{\textit{Budget: $K=50$ (Llama-3.2-1B)}} \\
CLT-RelP & $0.710 \pm 0.043$ & $0.501 \pm 0.052$ & $0.586 \pm 0.050$ \\
FActP & $0.723 \pm 0.042$ & $0.535 \pm 0.052$ & $0.614 \pm 0.048$ \\
FAP & $0.733 \pm 0.042$ & $0.540 \pm 0.051$ & $0.610 \pm 0.048$ \\
FAP-Synergy & $\mathbf{0.765} \pm 0.039$ & $\mathbf{0.581} \pm 0.049$ & $\mathbf{0.652} \pm 0.049$ \\
\midrule
\multicolumn{4}{l}{\textit{Budget: $K=100$ (Llama-3.2-1B)}} \\
CLT-RelP    & $0.596 \pm 0.043$ & $0.393 \pm 0.051$ & $0.467 \pm 0.053$ \\
FActP       & $0.617 \pm 0.045$ & $0.365 \pm 0.049$ & $0.451 \pm 0.051$ \\
FAP         & $0.631 \pm 0.045$ & $0.397 \pm 0.049$ & $0.486 \pm 0.050$ \\
FAP-Synergy & $\mathbf{0.637 \pm 0.046}$ & $\mathbf{0.399 \pm 0.049}$ & $\mathbf{0.488 \pm 0.050}$ \\
\midrule
\multicolumn{4}{l}{\textit{Budget: $K=50$ (Gemma-2-2B)}} \\
CLT-RelP    & $0.739 \pm 0.040$ & $0.631 \pm 0.045$ & $0.708 \pm 0.041$ \\
FActP       & $0.760 \pm 0.038$ & $0.659 \pm 0.043$ & $0.732 \pm 0.040$ \\
FAP         & $0.770 \pm 0.035$ & $0.644 \pm 0.046$ & $0.722 \pm 0.042$ \\
FAP-Synergy & $\mathbf{0.774 \pm 0.037}$ & $\mathbf{0.675 \pm 0.044}$ & $\mathbf{0.735 \pm 0.039}$ \\
\midrule
\multicolumn{4}{l}{\textit{Budget: $K=100$ (Gemma-2-2B)}} \\
CLT-RelP    & $0.596 \pm 0.042$ & $0.516 \pm 0.046$ & $\mathbf{0.597 \pm 0.044}$ \\
FActP       & $0.609 \pm 0.044$ & $0.502 \pm 0.045$ & $0.573 \pm 0.044$ \\
FAP         & $\mathbf{0.620 \pm 0.042}$ & $0.515 \pm 0.045$ & $0.588 \pm 0.044$ \\
FAP-Synergy & $0.619 \pm 0.042$ & $\mathbf{0.518 \pm 0.045}$ & $0.591 \pm 0.044$ \\
\bottomrule
\end{tabular}
\end{sc}
\end{small}
\end{table*}

\subsection{Case Study: Rescuing Synergistic Components}
\label{sec:case_study}

To illustrate the mechanism of FAP-Synergy, we analyze a specific "boundary" feature in Llama-3.2-1B that would have been discarded by standard magnitude pruning but was successfully rescued by our interaction-aware reranking.

We examine \textbf{Feature L0.2703}, interpreted as a "J-initial token detector" (activates on ``\textit{J}'', ``\textit{Java}'', ``\textit{Jacob}''). Under base FAP, this feature fell slightly below the Top-$K$ threshold ($S_{\text{FAP}}$ magnitude ranking) and was slated for removal. However, FAP-Synergy identified strong pairwise interactions with retained "Core" features, boosting its score to safe retention.

\paragraph{Positive Synergy: Constructive Amplification.}
Our analysis shows that L0.2703 provides crucial upstream support for higher-level semantic features. In particular, we identify \textbf{L5.8201} (\textit{``Given Name''} detector) as the \textbf{8th strongest positive-synergy partner} of L0.2703 (Synergy $\approx +0.156$): the model appears to use the low-level orthographic cue (\textit{``starts with J''}) from Layer 0 to amplify the confidence of the mid-level semantic signal (\textit{``is a name''}) at Layer 5. As a result, when both features are present, the model’s handling of names like \texttt{``Jacob''} exceeds what either feature achieves alone, whereas pruning the L0 trigger makes the L5 detector less reliable on this subset of names.

\paragraph{Negative Synergy: Managing Redundancy.}
Conversely, the framework also identified \textbf{Feature L0.5905} (specific "Jacob" detector) as the \textbf{1st highest negative synergy partner} ($Synergy \approx -0.375$). 
This negative interaction indicates redundancy: the specific \texttt{``Jacob''} feature and the general \texttt{``J-initial''} feature likely encode overlapping evidence for the token \texttt{``Jacob''}. This suggests either that L0.5905 is an overfitted backup of L0.2703 introduced during CLT reconstruction, or that activating L0.5905 creates a shortcut that suppresses L0.2703’s contribution in the Jacob context. By rescuing L0.2703, PIE preserves the circuit’s \textit{compositional} structure (Orthography $\rightarrow$ Semantics), rather than yielding a disjoint set of isolated semantic detectors.

\section{Discussion}
\label{sec:discussion}

\paragraph{PIE formalizes the prune-first benchmarking framework.}
A central obstacle in circuit analysis is deciding \emph{which} units warrant expensive auto-interpretation. PIE reframes this by establishing a rigorous, budgeted evaluation pipeline: prune candidates to a highly sparse causal core, then interpret. By adapting baselines like CLT-RelP and introducing FAP—which scales effortlessly via a single backward pass—PIE provides the community with a standardized benchmarking environment to measure the interpretability-fidelity tradeoff in CLTs, eliminating the need for exhaustive interpretation.

\paragraph{The sparsity gap justifies upstream pruning.}
As Figure~\ref{fig:fidelity_curve} demonstrates, random sampling from the \emph{active} feature set requires thousands of features to match the behavioral fidelity PIE achieves at just $K{=}100$. This massive $\approx$40$\times$ sparsity gap confirms that behaviorally relevant computation is concentrated in a tiny causal core, while the vast majority of active features are redundant. Isolating this core is scientifically necessary to sharpen mechanistic hypotheses, and economically essential to make automated interpretation viable.

\paragraph{Regime-specific recommendations and the value of synergy.}
Our benchmarking reveals distinct operational regimes. For highly relaxed sparsity constraints ($K \ge 400$), standard gradient and patching methods converge, making base FAP or CLT-RelP the computationally optimal choices. However, in strict-budget regimes ($K \le 100$) where downstream interpretation costs are highest, FAP-Synergy proves exceptionally advantageous. By functionally matching the fidelity of a $K{=}75$ base circuit with only $K{=}50$ features, synergy effectively grants the pipeline 25 ``free'' feature occurrences. This reduces downstream API costs by roughly 33\% while ensuring the retained set preserves coordinated, complementary computation rather than redundant backups.

\paragraph{Mechanistic lessons from the case study.}
The L0.2703 case study illustrates how PIE's synergy view supports \emph{mechanistic} claims rather than just aggregate metrics. First, it highlights cross-layer compositionality, showing that features with distinct roles (e.g., upstream orthographic cues and downstream semantic selection) must be preserved jointly. Second, it exposes the flaws of magnitude-only selection, which often retains highly active ``backup'' pathways that obscure true causal decompositions. By surfacing interaction partners, the synergy view confirms the extracted circuit is a coordinated functional subset, providing vital mechanistic validation.

\section{Conclusion}
We introduced PIE, the first CLT-native interpretability framework to formalize a ``prune first, interpret later'' paradigm. By connecting scalable pruning with targeted evaluation, PIE provides a rigorous benchmarking blueprint for mapping the internal mechanics of LLMs. We evaluated an adapted Relevance Patching baseline alongside our proposed Feature Attribution Patching (FAP) and FAP-Synergy, a systematic method that reranks near-threshold features. Empirically validating across multiple datasets and feature budgets using CLTs for Llama-3.2-1B and Gemma-2-2B, we identified clear regime-specific recommendations. At relaxed sparsity constraints ($K \ge 400$), standard gradient and patching methods converge, offering computationally efficient extraction. However, in most practical strict-budget regimes ($K \le 100$), FAP-Synergy proves highly advantageous. It yields a major ``Effective Budget'' advantage: by functionally matching the fidelity of a 75-feature-occurrence base circuit using only 50 features, it delivers target behavioral retention while reducing downstream auto-interpretation and evaluation costs by roughly 33\%. By establishing this evaluation standard, PIE offers a highly efficient, end-to-end foundation for future circuit discovery in CLTs.

\newpage

\bibliographystyle{plainnat} % or another natbib-compatible style like abbrvnat
\bibliography{example_paper}

\newpage
\appendix
\section{Hyperparameter Selection for FAP-Synergy}
\label{app:hyperparam}

\paragraph{Baseline.}
We use the ordinary FAP setting (synergy weight $\lambda=0$) as the baseline on IOI with $N{=}500$ (using a disjoint set from the 2000 IOI samples in the main experiment), $K{=}100$. The baseline metrics are: mean last-token KL $=1.1399$, std $=0.5393$, prediction-change rate $=0.446$. All results below are reported as \emph{deltas relative to this baseline}.

\paragraph{Selection rule.}
We select hyperparameters by \emph{minimizing mean last-token KL}. Even sub-percent improvements are meaningful in interpretability pruning studies, where improvements are typically incremental yet consistent across large prompt sets.

\subsection{Sensitivity Analysis}

We conducted a comprehensive sweep over the synergy weight $\lambda \in \{1, \dots, 5\}$ and boundary percent $bp \in \{20, \dots, 45\}$ to identify the optimal configuration for FAP-synergy. The results are summarized in Table~\ref{tab:sweep_delta} and visualized in Figure~\ref{fig:sweep_scatter}.

As shown in the table, the global optimum (lowest $\Delta$ mean KL) is achieved at $\lambda=3$ with a boundary percent of $bp=25$. The data reveals a consistent trend across all tested $\lambda$ values: increasing the boundary percent beyond $bp=25$ degrades performance (e.g., at $bp=40$, the improvement drops significantly to $\approx -0.758$ milli-KL compared to $> -1.0$ at $bp=25$). This suggests that widening the reranking window too far introduces noise or dilutes the high-synergy pairs with less relevant features. Furthermore, increasing $\lambda$ to 4 or 5 yields nearly identical results to $\lambda=3$ at the optimal $bp$, but does not surpass it. Consequently, we select $\lambda=3, bp=25$ as the most robust configuration.

\subsection{Delta table (vs.\ baseline)}
\begin{table}[h]
\centering
\caption{Sweep results reported as deltas vs.\ the $\lambda=0$ baseline (negative is better). 
$\Delta\mathrm{KL}$ and $\Delta\mathrm{std}$ are shown in \emph{milli-KL} (i.e., $\times 10^3$) for readability.
The selected setting is $\lambda=3$, bp$=25$.}
\label{tab:sweep_delta}
\scriptsize
\setlength{\tabcolsep}{6pt}
\begin{tabular}{cccc}
\toprule
$\lambda$ & bp & $\Delta$mean KL ($\times 10^3$) $\downarrow$ & $\Delta$std ($\times 10^3$) $\downarrow$ \\
\midrule
1 & 20 & -0.270 & -0.353 \\
1 & 25 & -0.785 & -0.770 \\
1 & 30 & -0.504 & -0.349 \\
1 & 35 & -0.668 & -0.509 \\
1 & 40 & -0.758 & -0.452 \\
1 & 45 & -0.203 & -0.165 \\
\midrule
2 & 20 & -0.074 & -0.293 \\
2 & 25 & -0.887 & -0.654 \\
2 & 30 & -0.488 & -0.334 \\
2 & 35 & -0.645 & -0.506 \\
2 & 40 & -0.758 & -0.452 \\
2 & 45 & -0.203 & -0.165 \\
\midrule
3 & 20 & -0.035 & -0.303 \\
\textbf{3} & \textbf{25} & \textbf{-1.066} & \textbf{-1.032} \\
3 & 30 & -0.504 & -0.339 \\
3 & 35 & -0.645 & -0.506 \\
3 & 40 & -0.758 & -0.452 \\
3 & 45 & -0.203 & -0.165 \\
\midrule
4 & 20 & -0.035 & -0.303 \\
4 & 25 & -1.043 & -1.015 \\
4 & 30 & -0.504 & -0.339 \\
4 & 35 & -0.645 & -0.506 \\
4 & 40 & -0.758 & -0.451 \\
4 & 45 & -0.203 & -0.165 \\
\midrule
5 & 20 & -0.027 & -0.303 \\
5 & 25 & -1.043 & -1.015 \\
5 & 30 & -0.488 & -0.323 \\
5 & 35 & -0.645 & -0.506 \\
5 & 40 & -0.758 & -0.451 \\
5 & 45 & -0.203 & -0.165 \\
\bottomrule
\end{tabular}
\end{table}

\subsection{Scatter plot (delta vs.\ baseline)}
\begin{figure}[h]
\centering
\begin{tikzpicture}
\begin{axis}[
    width=0.48\textwidth,
    height=0.35\textwidth,
    xlabel={boundary percent (bp)},
    ylabel={$\Delta$ mean last-token KL ($\times 10^3$)},
    ymin=-1.3, ymax=0.2,
    xtick={20,25,30,35,40,45},
    ytick distance=0.2,
    tick label style={font=\tiny},
    label style={font=\scriptsize},
    axis line style={black},
    tick style={black},
    grid=both,
    major grid style={line width=0.35pt, draw=gray!25},
    minor grid style={line width=0.2pt,  draw=gray!12},
    minor tick num=1,
    legend style={
        at={(0.5,-0.25)},
        anchor=north,
        legend columns=3,
        draw=none,
        fill=none,
        font=\tiny,
        row sep=2pt,
        /tikz/every even column/.append style={column sep=6pt},
    },
    legend cell align=left,
]

\addplot [black, thick] coordinates {(19.5,0) (45.5,0)};
\addlegendentry{baseline ($\lambda=0$)}

\addplot [
    only marks, mark=o, mark size=2.0pt,
    color=blue,
    mark options={solid, fill=white, draw=blue, line width=0.6pt}
] coordinates {
    (19.76,-0.270) (24.76,-0.785) (29.76,-0.504) (34.76,-0.668) (39.76,-0.758) (44.76,-0.203)
};
\addlegendentry{$\lambda=1$}

\addplot [
    only marks, mark=square*, mark size=2.0pt,
    color=orange,
    mark options={solid, fill=orange, draw=orange}
] coordinates {
    (19.92,-0.074) (24.92,-0.887) (29.92,-0.488) (34.92,-0.645) (39.92,-0.758) (44.92,-0.203)
};
\addlegendentry{$\lambda=2$}

\addplot [
    only marks, mark=triangle*, mark size=2.2pt,
    color=teal,
    mark options={solid, fill=teal, draw=teal}
] coordinates {
    (20.08,-0.035) (25.08,-1.066) (30.08,-0.504) (35.08,-0.645) (40.08,-0.758) (45.08,-0.203)
};
\addlegendentry{$\lambda=3$}

\addplot [
    only marks, mark=diamond*, mark size=2.2pt,
    color=violet,
    mark options={solid, fill=violet, draw=violet}
] coordinates {
    (20.24,-0.035) (25.24,-1.043) (30.24,-0.504) (35.24,-0.645) (40.24,-0.758) (45.24,-0.203)
};
\addlegendentry{$\lambda=4$}

\addplot [
    only marks, mark=x, mark size=2.2pt,
    color=red,
    mark options={solid, line width=0.9pt}
] coordinates {
    (20.40,-0.027) (25.40,-1.043) (30.40,-0.488) (35.40,-0.645) (40.40,-0.758) (45.40,-0.203)
};
\addlegendentry{$\lambda=5$}

\addplot [
    only marks, mark=star, mark size=3.5pt,
    color=black,
    mark options={fill=teal, draw=black, line width=0.9pt}
] coordinates {(25.08,-1.066)};
\addlegendentry{selected}

\end{axis}
\end{tikzpicture}
\caption{Sweep scatter plot reported as $\Delta$ mean KL vs. baseline ($\lambda=0$).}
\label{fig:sweep_scatter}
\end{figure}

\section{Implementation Details}
\label{app:implementation}

\subsection{Models and Resources}
We utilize the following public checkpoints:
\begin{itemize}
    \item \textbf{Gemma-2-2B CLT:} \texttt{mntss/clt-gemma-2-2b-426k} (Dictionary size: 426k).
    \item \textbf{Llama-3.2-1B CLT:} \texttt{mntss/clt-llama-3.2-1b-524k} (Dictionary size: 524k).
\end{itemize}
The IOI dataset is generated using the template structure from \citet{wang2023interpretability}, consisting of 2000 unique prompts with varying names and objects.

\subsection{Interpretation and Evaluation Protocol}

\paragraph{Step 2: Interpretation Generation.}
We use \texttt{gpt-5.2} as the explainer model.
\begin{itemize}
    \item \textbf{Exemplars:} We provide 40 max-activating examples per feature, retrieved from the Circuit Tracer activation cache.
    \item \textbf{Highlighting:} Tokens are highlighted if their activation value exceeds 65\% of the max activation in the sequence.
\end{itemize}

\paragraph{Step 3: FADE Evaluation.}
We use \texttt{gpt-5-mini} as the auditor model to compute the metrics defined in Section~\ref{sec:eval_pipeline}.
\begin{itemize}
    \item \textbf{Evaluation Dataset:} We draw samples from the \citep{sentence-transformers_wikipedia-en-sentences_2024} dataset.
    \item \textbf{Clarity:} The auditor generates 15 synthetic positive and negative examples based solely on the description. We measure the Gini coefficient of the feature's activations on these synthetic batches.
    \item \textbf{Purity and Responsiveness:} We evaluate the feature on a retrieved set of $N_{eval}=250$ real examples from Wikipedia (distinct from the exemplar set). 
\end{itemize}

\section{Pruning Dynamics and Budget Effects}
\label{app:stronger_baselines}

\paragraph{Feature Activation Patching behavior.}
In our feature-level Activation Patching adaptation, single-feature perturbations are informative only for a relatively small set of top features. We observe a low average number of non-zero activation patching effects per prompt (185.18 for Gemma-2-2B and 254.54 for Llama-3.2-1B), after which the ranking effectively degrades toward random selection. This is consistent with CLT redundancy and compensatory behavior under single-feature ablations. FAP is also substantially cheaper computationally, requiring one forward/backward pass rather than more than $4{,}000$ forward passes per prompt.

\paragraph{Cost-equivalent view of FAP-Synergy.}
Although the absolute KL differences between FAP and FAP-Synergy can appear small, they are meaningful in strict-budget regimes where interpretation cost scales linearly with the number of retained features. On IOI, base FAP at $K{=}75$ yields a KL of 1.22 on Llama-3.2-1B and 0.81 on Gemma-2-2B, while FAP-Synergy at $K{=}50$ achieves 1.22 and 0.82 respectively. Thus, in low-budget settings, FAP-Synergy can match the fidelity of a larger base-FAP circuit while reducing the downstream interpretation budget by roughly 33\%.

\paragraph{Failure mode under over-pruning.}
Very small budgets can remove features that carry important structural information even when they reduce computational cost. One illustrative case is \textbf{Feature L4.16331}, which functions as a conjunction detector linking two coordinated noun phrases, most often two person names. For the prompt, \textit{``Then, Arthur and Ruby had a long argument, and afterwards Ruby said to Arthur,''} this feature activates on the \textit{``and''} between \textit{Arthur} and \textit{Ruby}, providing a grammatical anchor for the coordinated-name structure. At $K{=}100$, both FAP and FAP-Synergy retain this feature, preserving this structural cue in the extracted circuit. At $K{=}50$, however, the feature is dropped entirely. The resulting circuit is still cheaper, but it loses a meaningful piece of grammatical structure, weakening the structural interpretability of the extracted mechanism. This example highlights a central risk of aggressive pruning: lower budgets may preserve coarse behavioral fidelity while discarding features that contribute important compositional or syntactic roles.

\section{Extended Behavioral Fidelity Results}
\label{sec:appendix_extended_fidelity}

Due to space constraints and our focus on highly sparse, human-interpretable circuits ($K \le 200$), the main text reports a truncated view of our pruning evaluations. Here, we present the exhaustive results for behavioral fidelity (Table~\ref{tab:kl_sweeps_full}) and faithfulness (Table~\ref{tab:faithfulness_sweeps_full}), extending the budget up to $K{=}800$.

\paragraph{Convergence at high budgets.} 
As observed in the tables below, the performance gap between CLT-RelP, base FAP, and FAP-Synergy narrows and eventually converges at $K \ge 400$. This validates the theoretical intuition that once the extracted budget is large enough to encompass the vast majority of the true causal graph, the precise boundary ranking provided by synergistic methods becomes less critical. If an interpretation pipeline allows for circuits of 500+ features, we suggest deploying base FAP or RelP for computational efficiency. 

\paragraph{Degradation of ACDC-style pruning.}
The extended tables also clearly illustrate the plateau and subsequent degradation of FActP. Because Cross-Layer Transcoders (CLTs) exhibit high robustness and redundancy to single-feature ablations, FActP yields zero effect for most features. Once the budget exceeds the small subset of non-zero effect features (averaging 185.18 for Gemma-2-2B and 254.54 for Llama-3.2-1B), FActP is forced to rank the remaining features randomly, causing its performance curve to heavily stagnate compared to gradient-based FAP methods.

\begin{table*}[ht]
\centering
\caption{\textbf{Comprehensive behavioral fidelity across tasks and extended budgets.} We report mean last-token KL (lower is better). While FAP-Synergy clearly dominates at $K \le 100$, gradient and patching methods converge as budgets grow large enough to encapsulate redundant features.}
\label{tab:kl_sweeps_full}
\small
\setlength{\tabcolsep}{3pt}
\begin{tabular}{llccccc}
\toprule
\textbf{Task / Model} & \textbf{Method} & \textbf{$K{=}50$} & \textbf{$K{=}100$} & \textbf{$K{=}200$} & \textbf{$K{=}400$} & \textbf{$K{=}800$} \\
\midrule
\multicolumn{7}{l}{\textit{IOI --- Llama-3.2-1B}} \\
& CLT-RelP & $1.32 \pm 0.62$ & $1.15 \pm 0.54$ & $0.87 \pm 0.41$ & $0.54 \pm 0.26$ & $0.22 \pm 0.11$ \\
& FAP & $1.33 \pm 0.60$ & $1.13 \pm 0.53$ & $0.85 \pm 0.42$ & $0.53 \pm 0.26$ & $0.22 \pm 0.13$ \\
& FAP-Synergy & $\mathbf{1.22 \pm 0.56}$ & $\mathbf{1.12 \pm 0.52}$ & $0.85 \pm 0.42$ & $0.53 \pm 0.26$ & $0.22 \pm 0.13$ \\
& Activation-Magnitude & 1.59 $\pm$ 0.69 & 1.52 $\pm$ 0.65 & 1.32 $\pm$ 0.59 & 1.08 $\pm$ 0.51 & 0.87 $\pm$ 0.43 \\
& FActP & $1.26 \pm 0.63$ & $1.24 \pm 0.64$ & $1.22 \pm 0.63$ & $1.19 \pm 0.61$ & $1.13 \pm 0.59$ \\
\midrule
\multicolumn{7}{l}{\textit{IOI --- Gemma-2-2B}} \\
& CLT-RelP & $0.86 \pm 0.43$ & $0.75 \pm 0.37$ & $0.53 \pm 0.28$ & $0.30 \pm 0.18$ & $0.14 \pm 0.09$ \\
& FAP & $0.91 \pm 0.46$ & $0.74 \pm 0.37$ & $0.52 \pm 0.29$ & $0.30 \pm 0.19$ & $0.15 \pm 0.10$ \\
& FAP-Synergy & $\mathbf{0.82 \pm 0.42}$ & $\mathbf{0.73 \pm 0.36}$ & $0.52 \pm 0.29$ & $0.30 \pm 0.19$ & $0.15 \pm 0.10$ \\
& Activation-Magnitude & 1.29 $\pm$ 0.59 & 1.25 $\pm$ 0.58 & 1.18 $\pm$ 0.56 & 0.97 $\pm$ 0.46 & 0.70 $\pm$ 0.33 \\
& FActP & $0.81 \pm 0.37$ & $0.77 \pm 0.38$ & $0.76 \pm 0.37$ & $0.72 \pm 0.33$ & $0.64 \pm 0.32$ \\
\midrule
\multicolumn{7}{l}{\textit{Doc-String --- Llama-3.2-1B}} \\
& CLT-RelP & $0.80 \pm 0.16$ & $0.72 \pm 0.15$ & $0.55 \pm 0.12$ & $0.35 \pm 0.08$ & $0.12 \pm 0.03$ \\
& FAP & $0.78 \pm 0.15$ & $0.69 \pm 0.14$ & $0.53 \pm 0.12$ & $0.32 \pm 0.08$ & $0.14 \pm 0.05$ \\
& FAP-Synergy & $\mathbf{0.74 \pm 0.14}$ & $\mathbf{0.68 \pm 0.14}$ & $0.53 \pm 0.12$ & $0.32 \pm 0.08$ & $0.14 \pm 0.05$ \\
& Activation-Magnitude & 0.88 $\pm$ 0.17 & 0.88 $\pm$ 0.17 & 0.85 $\pm$ 0.16 & 0.77 $\pm$ 0.14 & 0.68 $\pm$ 0.13 \\
& FActP & $0.78 \pm 0.16$ & $0.76 \pm 0.15$ & $0.76 \pm 0.15$ & $0.75 \pm 0.15$ & $0.74 \pm 0.15$ \\
\midrule
\multicolumn{7}{l}{\textit{Doc-String --- Gemma-2-2B}} \\
& CLT-RelP & $0.42 \pm 0.11$ & $0.37 \pm 0.10$ & $0.31 \pm 0.09$ & $0.20 \pm 0.07$ & $0.11 \pm 0.04$ \\
& FAP & $0.41 \pm 0.11$ & $0.36 \pm 0.10$ & $0.30 \pm 0.09$ & $0.21 \pm 0.07$ & $0.12 \pm 0.05$ \\
& FAP-Synergy & $\mathbf{0.40 \pm 0.11}$ & $0.36 \pm 0.10$ & $\mathbf{0.29 \pm 0.09}$ & $0.21 \pm 0.07$ & $0.12 \pm 0.05$ \\
& Activation-Magnitude & 0.43 $\pm$ 0.12 & 0.43 $\pm$ 0.12 & 0.43 $\pm$ 0.12 & 0.40 $\pm$ 0.12 & 0.36 $\pm$ 0.11 \\
& FActP & $0.40 \pm 0.12$ & $0.40 \pm 0.12$ & $0.39 \pm 0.12$ & $0.38 \pm 0.12$ & $0.36 \pm 0.11$ \\
\bottomrule
\end{tabular}
\end{table*}

\begin{table*}[ht]
\centering
\caption{\textbf{Comprehensive faithfulness on IOI across extended budgets.} We report mean faithfulness $\pm$ standard deviation. }
\label{tab:faithfulness_sweeps_full}
\small
\setlength{\tabcolsep}{5pt}
\begin{tabular}{llccccc}
\toprule
\textbf{Task / Model} & \textbf{Method} & \textbf{$K{=}50$} & \textbf{$K{=}100$} & \textbf{$K{=}200$} & \textbf{$K{=}400$} & \textbf{$K{=}800$} \\
\midrule
\multicolumn{7}{l}{\textit{IOI --- Llama-3.2-1B}} \\
& CLT-RelP & $0.15 \pm 0.17$ & $0.22 \pm 0.20$ & $0.37 \pm 0.22$ & $0.57 \pm 0.22$ & $0.75 \pm 0.15$ \\
& FAP  & $0.27 \pm 0.17$ & $0.34 \pm 0.17$ & $0.43 \pm 0.17$ & $0.58 \pm 0.18$ & $0.76 \pm 0.16$ \\
& FAP-Synergy  & $0.30 \pm 0.16$ & $0.35 \pm 0.15$ & $0.43 \pm 0.16$ & $0.58 \pm 0.18$ & $0.76 \pm 0.16$ \\
\midrule
\multicolumn{7}{l}{\textit{IOI --- Gemma-2-2B}} \\
& CLT-RelP & $0.26 \pm 0.45$ & $0.35 \pm 0.60$ & $0.54 \pm 0.54$ & $0.75 \pm 0.73$ & $0.96 \pm 0.77$ \\
& FAP  & $0.29 \pm 0.38$ & $0.39 \pm 0.43$ & $0.55 \pm 0.46$ & $0.74 \pm 0.51$ & $0.90 \pm 0.52$ \\
& FAP-Synergy  & $0.34 \pm 0.40$ & $0.40 \pm 0.44$ & $0.55 \pm 0.46$ & $0.74 \pm 0.51$ & $0.90 \pm 0.52$ \\
\bottomrule
\end{tabular}
\end{table*}

\section{Economic Analysis}
\label{app:eco_analysis}

Our PIE pipeline incurs API cost only in the \emph{Interpret} and \emph{Evaluate} stages. Under our experimental configuration, the \textbf{explanation} phase (GPT-5.2) consumes approximately $4{,}000$ input tokens (system prompt + 40 max-activating exemplars) and produces $\approx 200$ output tokens \emph{per feature}. The subsequent \textbf{evaluation} phase (GPT-5 mini, used for both synthetic clarity generation and purity rating) is more data-intensive, averaging $22{,}650$ input tokens and $4{,}000$ output tokens per feature. Under standard pricing\citep {openai_api_pricing_2026}, this yields a total estimated cost of
\[
c_{\text{feat}} \approx \$0.0235 \quad \text{per interpreted feature.} 
\]

\paragraph{Prompt-level budgeting (active set vs.\ $K{=}100$).}
A rigorous alternative to interpreting \emph{all} CLT features is to interpret only those that are \emph{active} for the prompt. In our setting, the mean number of active feature occurrences per prompt is $4{,}188$ for Llama-3.2-1B and $5{,}190$ for Gemma-2-2B.

Interpreting the full active set for a \emph{single} prompt would therefore cost
\begin{align}
\text{Cost}_{\text{Llama, active/prompt}} &\approx 4{,}188 \cdot c_{\text{feat}} \approx \$98.42,\\
\text{Cost}_{\text{Gemma, active/prompt}} &\approx 5{,}190 \cdot c_{\text{feat}} \approx \$121.97.
\end{align}
In contrast, PIE applies a strict per-prompt budget of $K=100$ occurrences. Because this yields at most 100 unique features per prompt, the maximum amortized interpretation cost for a single prompt is:
$$Cost_{K=100, prompt} \le 100 \cdot c_{feat} \approx \$2.35$$
Thus, even when comparing against the strong baseline that restricts attention to active features, PIE reduces the per-prompt interpretation spend by a factor of $\frac{4,188}{100} \approx 41.9\times$ (Llama) and $\frac{5,190}{100} \approx 51.9\times$ (Gemma), directly mirroring the causal sparsity gap.

\paragraph{Dataset-level budgeting (global reuse across 2,000 prompts).}
The prompt-level view is conservative because it does not exploit reuse: across a dataset, the same features appear repeatedly, so we can cache interpretations and only pay once per \emph{unique} feature. Concretely, when aggregating across $N{=}2000$ prompts, the number of \emph{unique} features that ever appear as \texttt{kept} after pruning is approximately $4{,}400$ for Llama and $4{,}000$ for Gemma. Under this global accounting, the total interpretation+evaluation cost becomes
\begin{align}
\text{Cost}_{\text{Llama, global kept}} &\approx 4{,}400 \cdot c_{\text{feat}} \approx \$103.40,\\
\text{Cost}_{\text{Gemma, global kept}} &\approx 4{,}000 \cdot c_{\text{feat}} \approx \$94.00.
\end{align}
By comparison, a naive global sweep that attempts to interpret the \emph{entire} CLT dictionary would require evaluating
\begin{equation}
\begin{split}
|\mathcal{F}_{\text{Llama}}| &= 16 \times 32{,}768 = 524{,}288, \\
|\mathcal{F}_{\text{Gemma}}| &= 26 \times 16{,}384 = 425{,}984.
\end{split}
\end{equation}
which would cost
\begin{align}
\text{Cost}_{\text{Llama, full dict}} &\approx 524{,}288 \cdot c_{\text{feat}} \approx \$12{,}320.77,\\
\text{Cost}_{\text{Gemma, full dict}} &\approx 425{,}984 \cdot c_{\text{feat}} \approx \$10{,}010.62.
\end{align}
Therefore, from a global perspective PIE reduces the evaluation burden by $\frac{524{,}288}{4{,}400}\approx 119\times$ on Llama and $\frac{425{,}984}{4{,}000}\approx 106\times$ on Gemma, translating to savings on the order of ${\sim}\$12.2\text{k}$ and ${\sim}\$9.9\text{k}$ for a single end-to-end run.

\begin{table*}[t]
\centering
\small
\begin{tabular}{lrrrr}
\toprule
\textbf{Scenario} & \textbf{\#Features} & \textbf{Cost} & \textbf{vs.\ Budgeted} & \textbf{Reduction} \\
\midrule
\multicolumn{5}{c}{\textit{Prompt-level (single prompt)}}\\
Llama active set & 4{,}188 & \$98.42 & \$2.35 ($K{=}100$) & $41.9\times$ \\
Gemma active set & 5{,}190 & \$121.97 & \$2.35 ($K{=}100$) & $51.9\times$ \\
\midrule
\multicolumn{5}{c}{\textit{Global (2,000 prompts; unique features)}}\\
Llama kept (unique) & 4{,}400 & \$103.40 & full dict: \$12{,}320.77 & $119\times$ \\
Gemma kept (unique) & 4{,}000 & \$94.00 & full dict: \$10{,}010.62 & $106\times$ \\
\bottomrule
\end{tabular}
\caption{\textbf{Economic impact of budgeting.} Using $c_{\text{feat}}\approx \$0.0235$ per feature, a fixed interpretation budget (and global reuse across prompts) yields large cost reductions relative to interpreting the full active set per prompt or sweeping the entire CLT dictionary.}
\label{tab:econ}
\end{table*}

\section{Interpretability metrics on IOI tasks}
\label{app:fade_ioi}

In Table~\ref{tab:interpretability_results_ioi}, we present the comprehensive interpretability quality metrics for the Indirect Object Identification (IOI) task at a strict budget of $K=50$. As noted in Section 4.2, space limitations precluded the inclusion of these full results in the main text.

Consistent with the Docstring evaluation, the IOI results demonstrate that FAP-Synergy maintains a distinct advantage over standard relevance-based (CLT-RelP) and factorization-based (FActP) methods. Specifically, FAP-Synergy achieves the highest Clarity and Purity scores for both the Llama-3.2-1B and Gemma-2-2B models. While CLT-RelP remains highly competitive on the Responsiveness metric for Llama-3.2-1B, FAP-Synergy provides the most robust overall balance across the three metrics. These supplementary findings reinforce the conclusion that accounting for boundary interactions during pruning systematically rescues features that are highly interpretable and semantically coherent.

\begin{table*}[ht]
\centering
\caption{Interpretability quality of pruned circuits at $K=50$ budgets on IOI tasks. We report Mean $\pm$ Std for quality metrics. FAP-Synergy maintains the highest overall interpretability.}
\label{tab:interpretability_results_ioi}
\begin{small}
\begin{sc}
\begin{tabular}{lccc}
\toprule
\textbf{Method} & \textbf{Clarity}  & \textbf{Purity}  & \textbf{Responsiveness}  \\
\midrule
\multicolumn{4}{l}{\textit{Budget: $K=50$ (Llama-3.2-1B)}} \\
CLT-RelP    & $0.722 \pm 0.042$ & $0.608 \pm 0.050$ & $\mathbf{0.685} \pm 0.045$ \\
FActP       & $0.735 \pm 0.042$ & $0.591 \pm 0.048$ & $0.649 \pm 0.046$ \\
FAP         & $0.742 \pm 0.041$ & $0.573 \pm 0.050$ & $0.639 \pm 0.048$ \\
FAP-Synergy & $\mathbf{0.770} \pm 0.039$ & $\mathbf{0.622} \pm 0.047$ & $0.682 \pm 0.044$ \\
\midrule

\multicolumn{4}{l}{\textit{Budget: $K=50$ (Gemma-2-2B)}} \\
CLT-RelP    & $0.733 \pm 0.043$ & $0.691 \pm 0.040$ & $0.756 \pm 0.037$ \\
FActP       & $0.734 \pm 0.040$ & $0.676 \pm 0.040$ & $0.737 \pm 0.038$ \\
FAP         & $0.745 \pm 0.039$ & $0.670 \pm 0.043$ & $0.731 \pm 0.040$ \\
FAP-Synergy & $\mathbf{0.769} \pm 0.038$ & $\mathbf{0.699} \pm 0.040$ & $\mathbf{0.757} \pm 0.038$ \\

\bottomrule
\end{tabular}
\end{sc}
\end{small}
\end{table*}

\section{Limitations}
\label{app:limitations}

Our results should be interpreted in light of several practical limitations.

\paragraph{Synergy may be under-measured by ``feature-isolated'' evaluation.}
The case study suggests that synergy-aware reranking's effect is \emph{structural}: it preferentially retains \emph{complementary} features whose joint presence preserves coordinated computation, rather than substitutable ``backup'' features that match behavior only marginally. In other words, synergy helps preserve \emph{interactions}, in which the way features compose into a circuit, even when their marginal, feature-by-feature contributions look similar at the pruning boundary. This distinction may be weakly expressed in our present evaluation setting because our interpretability pipeline largely treats features as \emph{isolated} units (explained and scored independently), which is not fully ``circuit-friendly'' and can under-reward interaction-preserving selections. The mechanistic lessons therefore motivate an important direction for future work: developing a \emph{CLT-native} evaluation pipeline that explicitly evaluates \emph{sets} of features and their composition (e.g., interaction-aware interventions, group-level counterfactuals, and structure-sensitive explanation scoring). Such an evaluation would better capture the value of synergy-preserving selection and enable a clearer understanding of when and why synergy improves circuit faithfulness beyond what marginal metrics reveal.

\paragraph{Interpretability evaluation remains concentrated at $K{=}100$.}
We report broader behavioral-fidelity sweeps over $K\in\{50,100,200,400,800\}$ and across two tasks, but our full end-to-end interpretation evaluation (including FADE-style metrics and economic analysis) remains centered on the main $K{=}100$ setting. As a result, while the pruning robustness story now extends beyond a single budget, we do not yet report a full scaling study of downstream interpretability quality across the entire budget range.

\paragraph{No CLT training; reliance on public replacement checkpoints.}
We do not train Cross-Layer Transcoders (CLTs) ourselves; instead, we use public CLT checkpoints released with Circuit Tracer tooling (Gemma-2-2B CLT and Llama-3.2-1B CLT). Consequently, our claims are conditional on the quality and representational coverage of these replacement models. In particular, pruning behavior and feature semantics may differ for CLTs trained with different data, objectives, sparsity regimes, or architectures.

\paragraph{Scope of causal claims.}
PIE is designed to isolate a minimal circuit that reproduces behavior under the replacement-model intervention, but this does not automatically imply that every retained feature corresponds to a unique mechanistic ``part'' in a human-interpretable decomposition. Feature redundancy, polysemanticity, and interaction effects can persist even after pruning, and the replacement-model abstraction may miss mechanisms not captured by the CLT basis.

\section{Impact Statement}
\label{app:impact statement}
Our work aims to make mechanistic interpretability scalable and economically viable. By enabling the pruning of Cross-Layer Transcoder (CLT) features, the PIE framework significantly reduces the computational and financial costs associated with automated model interpretation. This advances the goal of AI transparency, making it feasible to audit large models for safety-critical behaviors without analyzing millions of redundant parameters. Furthermore, by lowering the resource requirements for circuit discovery, this work promotes inclusivity in the research community, allowing entities with smaller compute budgets to participate in safety research. However, we acknowledge the risk of over-reliance on pruned circuits; aggressive pruning might obscure subtle, distributed computations necessary for full behavioral robustness. Researchers should treat pruned interpretations as high-signal approximations rather than exhaustive descriptions of model psychology.
\section{LLM Usage Acknowledgment}
\label{app:llm usage}
The authors acknowledge the use of Large Language Models (LLMs) to assist in the preparation of this manuscript. Specifically, LLMs were utilized to improve the clarity and flow of the writing, assist in filtering and parsing structured data, and aid in the generation of data visualizations. All core concepts, experimental designs, and interpretations remain the original work and responsibility of the authors.
\section{Computing Resources Usage}
\label{app:computing_resources}

\paragraph{Hardware Specifications}
All local feature extraction, patching, and pruning experiments were executed on a single NVIDIA A40 GPU.

\paragraph{Computational Cost for Circuit Discovery}
The processing time for evaluating a dataset of 2000 prompts varies significantly depending on the selected feature selection method and the target budget constraint ($K$). 

\begin{itemize}
    \item \textbf{Gradient and Relevance Methods (Low-Budget Regimes):} For strict-budget settings ($K \le 200$), calculating the pruning scores and extracting the circuits using Feature Attribution Patching (FAP), FAP-Synergy, Activation-Magnitude, and CLT-RelP requires approximately 24 hours per dataset on our hardware setup.
    \item \textbf{Synergy-Aware Scaling:} Because FAP-Synergy systematically evaluates pairwise interactions at the pruning boundary to rescue complementary features, its computational overhead scales with the budget size. While it operates within the standard 24-hour window for constrained, interpretable regimes, evaluating a highly relaxed budget of $K=800$ extends the processing time to approximately 48 hours.
    \item \textbf{Perturbation-Based Baseline (FActP):} Iterative, single-feature ablation methods exhibit prohibitive computational scaling. Running FActP over 2000 prompts requires roughly 200 compute hours on a single A40. This massive disparity further motivates the necessity of gradient-weighted proxies like FAP for scalable circuit discovery.
\end{itemize}

\section{Licenses for Existing Assets}
\label{app:licenses}
We acknowledge the creators of the foundational models utilized in this research. The \textbf{Gemma-2} model (powering our Gemma-2-2B CLT) is used in accordance with the \href{https://ai.google.dev/gemma/terms}{Gemma Terms of Use}. The \textbf{Llama-3.2} model is licensed under the Llama 3.2 Community License, Copyright \copyright~Meta Platforms, Inc. All Rights Reserved.

\end{document}